\documentclass[conference]{IEEEtran}
\usepackage{blindtext, graphicx}
\usepackage{multirow}

\ifCLASSINFOpdf
 \else
\fi

\begin{document}
\title{A Deep Learning Framework using Passive WiFi Sensing for Respiration Monitoring}
\author{\IEEEauthorblockN{Usman Mahmood Khan\IEEEauthorrefmark{1},
Zain Kabir\IEEEauthorrefmark{1},
Syed Ali Hassan\IEEEauthorrefmark{1},\\ 
Syed Hassan Ahmed\IEEEauthorrefmark{2}}
\IEEEauthorblockA{\IEEEauthorrefmark{1}School of Electrical Engineering \& Computer Science (SEECS), National University of Sciences \& Technology\\
(NUST), Islamabad, Pakistan \{13beeukhan, 13beezkabir, ali.hassan\}@seecs.nust.edu.pk}
\IEEEauthorblockA{\IEEEauthorrefmark{2}Kyungpook National University \\ s.h.ahmed@ieee.org}
}

% make the title area
\maketitle

\begin{abstract}
\noindent This paper presents an end-to-end deep learning framework using passive WiFi sensing to classify and estimate human respiration activity. A passive radar test-bed is used with two channels where the first channel provides the reference WiFi signal, whereas the other channel provides a surveillance signal that contains reflections from the human target. Adaptive filtering is performed to make the surveillance signal source-data invariant by eliminating the echoes of the direct transmitted signal. We propose a novel convolutional neural network to classify the complex time series data and determine if it corresponds to a breathing activity, followed by a random forest estimator to determine breathing rate. We collect an extensive dataset to train the learning models and develop reference benchmarks for the future studies in the field. Based on the results, we conclude that deep learning techniques coupled with passive radars offer great potential for end-to-end human activity recognition.

\end{abstract}


\IEEEpeerreviewmaketitle

\section{Introduction}

\noindent 

\noindent Ubiquitous health monitoring has emerged as a key area of interest over the past few years \cite{1} \cite{2} . Frequently, the monitoring is targeted at tracking vital health signs such as the human breathing in a non-intrusive manner. Human breathing involves continuous inhale and exhale movements and may be used to study the subject's physiological state, her stress levels, or even emotions like \textit{fear} and \textit{relief} \cite{3}. A common trend in this field is to use wearable devices \cite{4} but they present several challenges, e.g., they intrude with users' routine activities, they have to be worn all the time even during sleep, and they have to be frequently recharged. These factors and others make wearable devices unattractive in practice.

With the advances in wireless sensing, it has become possible to use non-contact devices to determine signs of life \cite{5} \cite{6}. Such devices can be broadly classified into two categories depending on the kind of radar sensing technique they use. Active radars employ dedicated transmitters, use high bandwidth and often require complex antenna arrays to go along with them \cite{7}. In contrast, passive radars are covert, require low bandwidth and utilize the illuminations of opportunity in the environment such as WiFi and cellular signals for target monitoring \cite{8}. Despite the advantages, passive radars typically require complex signal processing and feature engineering to accurately track activities. This gives rise to the question: can we leverage deep learning techniques to automatically develop feature representations from the sensed data?

A \textit{deep learning} network uses a cascade of multiple layers of non-linear processing units, where each successive layer takes the output from the previous layer as an input \cite{9}. At each abstraction level, higher order features are constructed from the lower level features from the previous layers \cite{10}. To explain the motivation for a deep learning model, consider a typical radio communication system that uses \textit{matched filter} receivers to identify the correct transmitted symbols. The intuition behind the deep learning model is that during training, the model will learn to form matched filters or representations of data to recognize various activities. The convolutional neural network (CNN) is a deep learning model that uses spatial information between the data samples to form learned representations of data \cite{11}. Since convolutional networks have shared weights \cite{12}, translation invariance is achieved which means that only the salient features in the data are considered and not their positions or scales.

\begin{figure}[!t]
\centering
\includegraphics[scale=0.7]{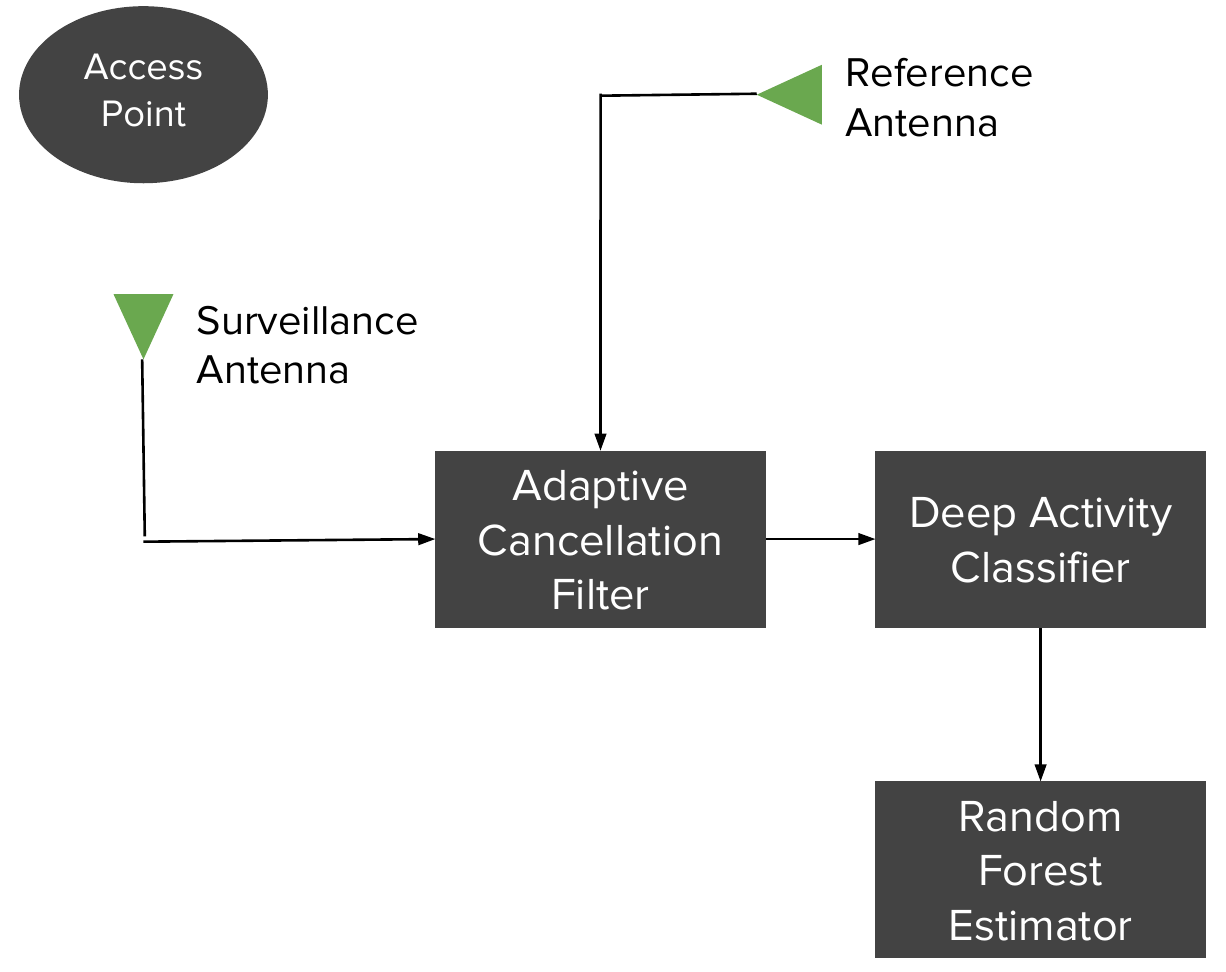}
\caption{System block model diagram with access point, surveillance and reference antennas, and signal processing algorithms.}
\end{figure} 

Fig. 1 provides a high level overview of the system which is explained as follows. A WiFi access point serves as the illuminator of opportunity and two signals, a reference signal and a surveillance signal (reflected from target), are observed. An adaptive filter is applied to the received signal at the surveillance antenna in order to remove the unwanted echoe of the direct transmitted signal. The data after the cancellation is fed to a deep learning network that classifies whether or not a movement corresponds to the breathing activity. Depending on the outcome of this stage, a Random Forest estimator is used to determine the breathing rate. 

To the best of the authors' knowledge, this is the first comprehensive study on deep learning techniques using passive WiFi sensing for basic activity classification. Specifically, this research introduces a vital health wireless device and makes the following contributions: 

\begin{itemize}
\item Proposes a deep learning framework for classifying various activities using passive WiFi sensing

\item Creates a test-bed with software defined radios to carry out deep learning experiments

\item Collects an extensive human activity dataset that can be used as a benchmark in future studies

\item Carries out experiments to determine breathing classification and breathing estimation accuracy
\end{itemize}

\subsection{Related Work}
Multiple approaches have been proposed for human motion detection and other applications using passive sensing. Kotaru \textit{et.al} presented an indoor localization system using the channel state information (CSI) and received signal strength information (RSSI) \cite{13}. Similar approaches have been used for other applications such as keystroke identification \cite{14}, in-home activity analysis \cite{15}, and virtual drawing \cite{16}. Some work has been done on using active radars for breathing rate measurements. WiZ uses frequency modulated continuous wave (FMCW) signal chirps to detect breathing rate in both line-of-sight (LOS) and non line-of-sight (NLOS) scenarios \cite{17}. Other similar works include an ultra wide-band radar with short transmit pulses to avoid reflections from surrounding objects \cite{18}, a continuous wave (CW) radar for clinical measurements \cite{19}, and a CW radar for indoor environment \cite{3}. 

Deep learning techniques using passive WiFi sensing are mostly unexplored. Yang \textit{et.al} proposed a deep CNN on multichannel time series for human activity recongition using the on-body sensors \cite{10}. With the rise of deep learning techniques in the computer vision domain \cite{20} \cite{21}, various radio applications have also begun to capture interest of deep learning researchers. O'shea \textit{et.al} proposed CNNs for signal modulation recongition \cite{22}. Other researchers have targeted applications like solar radio burst classification \cite{9}, and CSI-based fingerprinting for indoor localization \cite{11}.

The rest of the paper is organized as follows. Section II provides an exhaustive system overview and covers the passive radar system, adaptive filtering, and deep learning stages. In Section 3, a detailed implementation of the machine learning networks is covered, followed by experimental details in Section IV. Section V lists the key results and limitations of this research, and gives insights into the future research directions.

\section{System Overview}

\subsection{Passive Radar System}
Passive radars use signal of opportunity to track motion changes in the environment \cite{8}. Fig. 2 shows a passive radar system with an access point (AP), a reference antenna and a surveillance antenna. The reference antenna collects the direct signal from the AP, which is later used in the adaptive cancellation step. The surveillance antenna tracks both the transmitted signal and the variations caused due to the human breathing movement, and can be represented as

\begin{figure}[!t]
\centering
\includegraphics[scale=0.46]{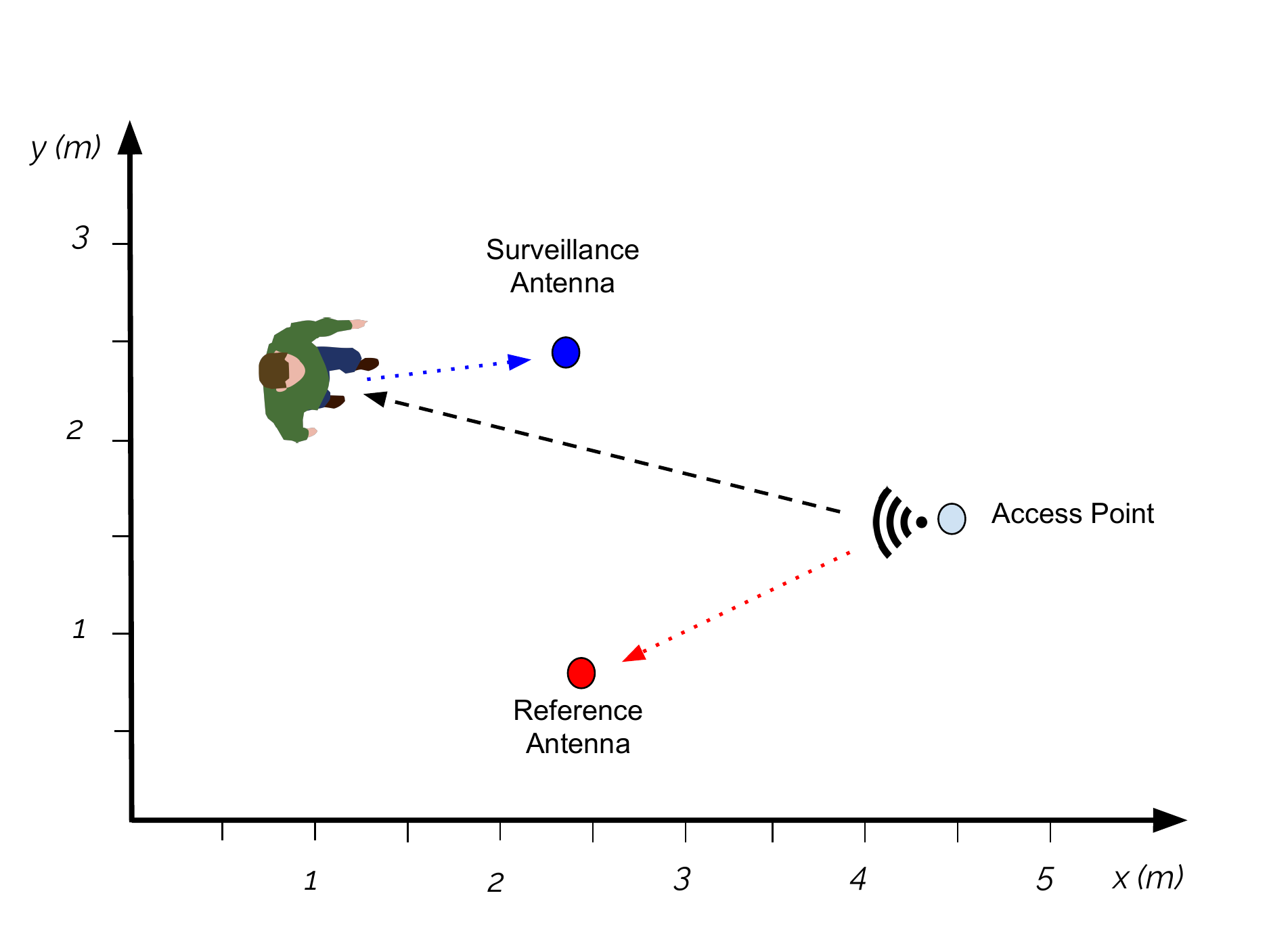}
\caption{Passive radar system with an access point, a surveillance antenna and a reference antenna.}
\end{figure}

\begin{equation}
{s[n] = \sum_{f_d \in f_D} Ax[n+\tau]e^{i2\pi f_{d}n}},
\end{equation}

\noindent where \(x[n]\) is the source signal, \(\tau\) is the path delay from AP to the surveillance receiver, \(A\) is the signal amplitude, and \(f_D\) contains a set of Doppler shifts in the surveillance antenna plane. In the simplest case, when there is a single reflection at the surveillance antenna from the desired breathing movement, the surveillance signal equation simplifies to

\begin{equation}
{s[n] = Ax[n+\tau]e^{i2\pi f_{d}n}}.
\end{equation}

In the given passive radar system, the source signal \(x[n]\) is not constant. In order to train the system to recongize breathing patterns, the surveillance signal must be modified in a way such that it is insensitive to the changes in \(x[n]\). In other words, the modified surveillance signal should be source data invariant and only vary with changes in signal amplitude or associated Doppler shifts. The following section explains an adaptive filtering approach to modify the surveillance signal in such a desired manner.

\subsection{Adaptive Cancellation}

Let the reference signal is given by \(r[n]\). The goal of the adaptive canceller is to eliminate the direct echo of the transmitted signal from the surveillance signal commonly known in literature as direct signal attenuation (DSA). The adaptive filter uses a least mean square (LMS) update algorithm, which updates the filter output \(y[n]\) based on the following rule

\begin{equation}
{y[n] = w' \times r[n]},
\end{equation}

\noindent where \((.)'\) denotes the conjugate operator, \(w\) defines the weight of the filter, and \(r[n]\) represents the reference signal. The error signal of the adaptive filter is calculated by subtracting the filter output from the surveillance signal, and denotes the true echo signal of interest. At each iteration, the filter weight coefficient is updated as follows

\begin{equation}
{w = w + \mu \times s[n] \times e[n]'},
\end{equation}

\noindent where \(\mu\) is the step-size parameter that controls how quickly the LMS algorithm converges \cite{23}. Through this process, the surveillance signal \(s[n]\) adjusts its weights so that the output of the filter closely corresponds to the reference signal \(r[n]\) and the error signal \(e[n]\) is fed to the later stages for further processing.

\subsection{Deep Activity Classification}

\begin{figure}[!t]
\centering
\includegraphics[scale=0.44]{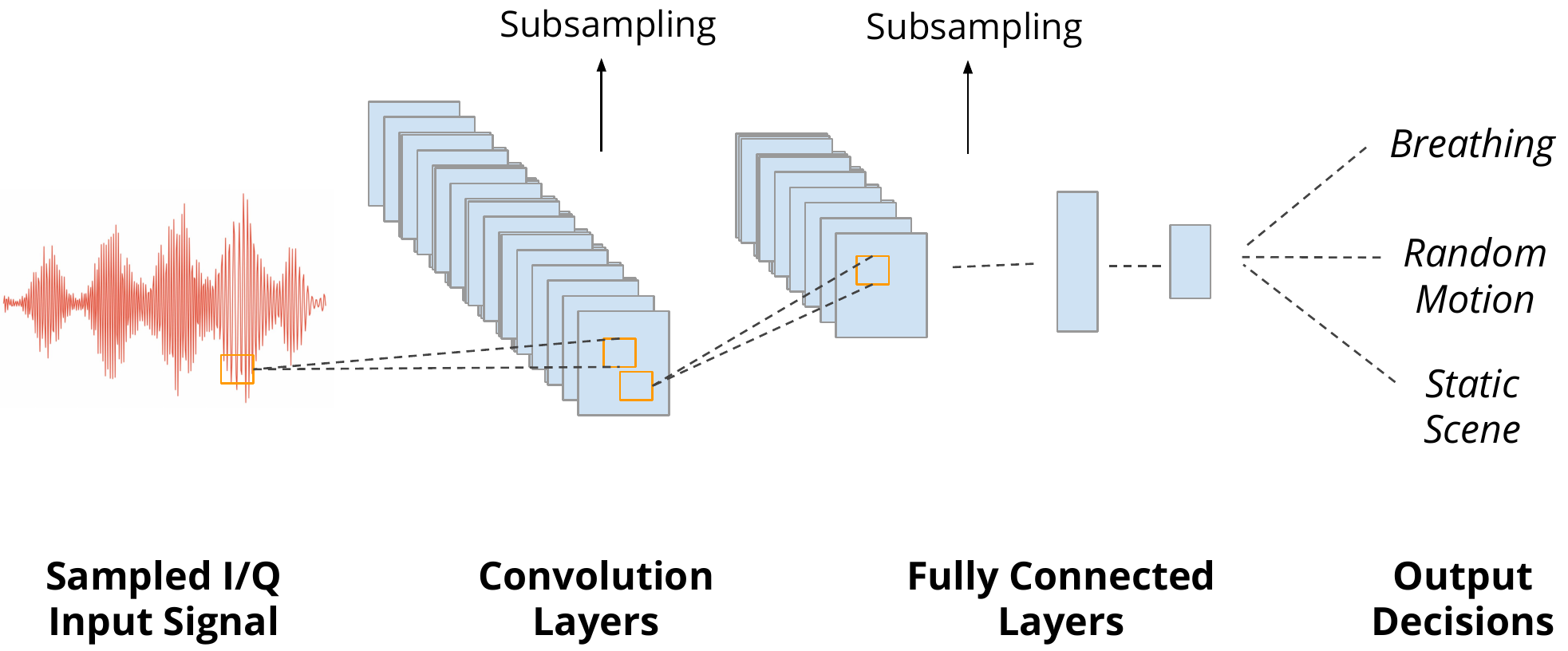}
\caption{CNN architecture for activity classification. The sampled I/Q error signal is fed to convolution layers which act as matched filters, followed by a set of fully connected layers. }
\end{figure}

Before applying the deep learning model, the raw time series error signal obtained from the previous adaptive filtering stage is segmented into a collection of short pieces of signals using a sliding window strategy. After segmentation, the dimension of the time series signal is given by \(2 \times R\), where \(R\) is the number of samples in a single window. Fig. 3 shows the CNN design for breathing rate recongition task where sampled I/Q signal is fed first to convolution layers and then to fully connected layers. In the convolution layers, feature maps from the previous layers are convolved with several convolution kernels with a depth \(D\). Afterwards, a bias value is added and an activation function is applied to the output to form output feature maps for the next layer. Concretely, the \(m^{th}\) feature map in the \(n^{th}\) layer of the CNN is written as \(v_{mn}^{rc}\), where \(r\) is the data row and \(c\) denotes either real or imaginary channel. Formally, \(v_{mn}^{rc}\) is represented as

\begin{equation}
{v_{mn}^{rc} = max (0, (b_{mn} + \sum_{m}\sum_{l=0}^{L_i - 1} k_{mni}^{p} v_{(m-1)i}^{(r+l),c}))},
\end{equation}

\noindent where \(max(0, x)\) denotes the rectified linear unit activation function (ReLU), \(b_{mn}\) is the bias value of the current feature map, \(m\) is the index of feature maps in the \((i - 1)^{th}\) layer, \(k_{mni}^{p}\) is the current kernel value, and \(L\) is the kernel length. The fully connected layer (FCL) transforms the feature maps into the output classes. In order to increase feature invariance to input distortions, a pooling layer is used where the resolution of feature maps is reduced. Specifically, the feature maps from the convolutional layer are pooled over a local temporal vicinity by the max pooling function which is given as

\begin{equation}
{v_{mn}^{rc} = \max_{1 \leq h \leq H_i}{(v_{(m-1)n}^{(r+q)c})}},
\end{equation}

\noindent where \(d\) is equal to \(\{1, ... , D\}\) and \(H_i\) is the length of the pooling region. The output of the FCL is governed by the softmax function, i.e.,

\begin{equation}
{v_{mn}^{rc} = \frac{exp(v_{(m-1)n}^{rc})}{\sum_{j=1}^{C}exp(v_{(m-1)j}^{rc})}},
\end{equation}

\noindent where \(C\) is the number of output classes. The overall model can be trained by using an entropy cost function which is based on the true data labels and the probabilistic outcomes generated through the above softmax function \cite{24}. In order to evaluate the model's effectiveness, the following metrics are defined

\begin{equation}
{Accuracy = \frac{TP + TN}{TP + TN + FP + FN}},
\end{equation} 

\noindent where \(TP\) is the number of true positives, \(TN\) is the number of true negatives, \(FP\) is the number of false positives and \(FN\) is the number of false negatives. The other metrics are sensitivity and specificity and are defined as

\begin{equation}
{Sensitivity = \frac{TP}{TP + FN}},
{Specificity = \frac{FP}{FP + TN}}.
\end{equation} 

\subsection{Breathing Rate Estimation}
Once the breathing activity is identified, features are extracted from the signal and breathing rate estimation is done through the Random Forest algorithm. In general, the \(n^{th}\) feature corresponding to the \(m^{th}\) operator is given by

\begin{equation}
{f_{nm} = S_{n}{(O_{m}{(e[n])})}},
\end{equation}

\noindent where \(S_{n}\) is the \(n^{th}\) order statistical moment and \(O_{m}\) is a mathematical operator such as real valued operator on a complex number. Random Forest differs from a traditional decision tree in the respect that it trains multiple trees on various parts of the feature space, ensuring that no tree is exposed to the entire training data. The data is recursively partitioned and each split is made using Gini impurity criteria \cite{25} which is given at node \(N\) by

\begin{equation}
{g(N) = N_l \sum_{k=1}^{K}p_{kl}(1-p_{kl}) + N_r \sum_{k=1}^{K}p_{kr}(1-p_{kr})},
\end{equation}

\noindent where \(p_{kl}\) is the proportion of class \(k\) in the child node left to \(N\), \(p_{kr}\) is the proportion of class \(k\) in the right node, and \(N_l\) and \(N_r\) are the left and right children of \(N\), respectively \cite{25}.

\subsection{System Range and Limitations}
\begin{figure}[!t]
\centering
\includegraphics[scale=0.42]{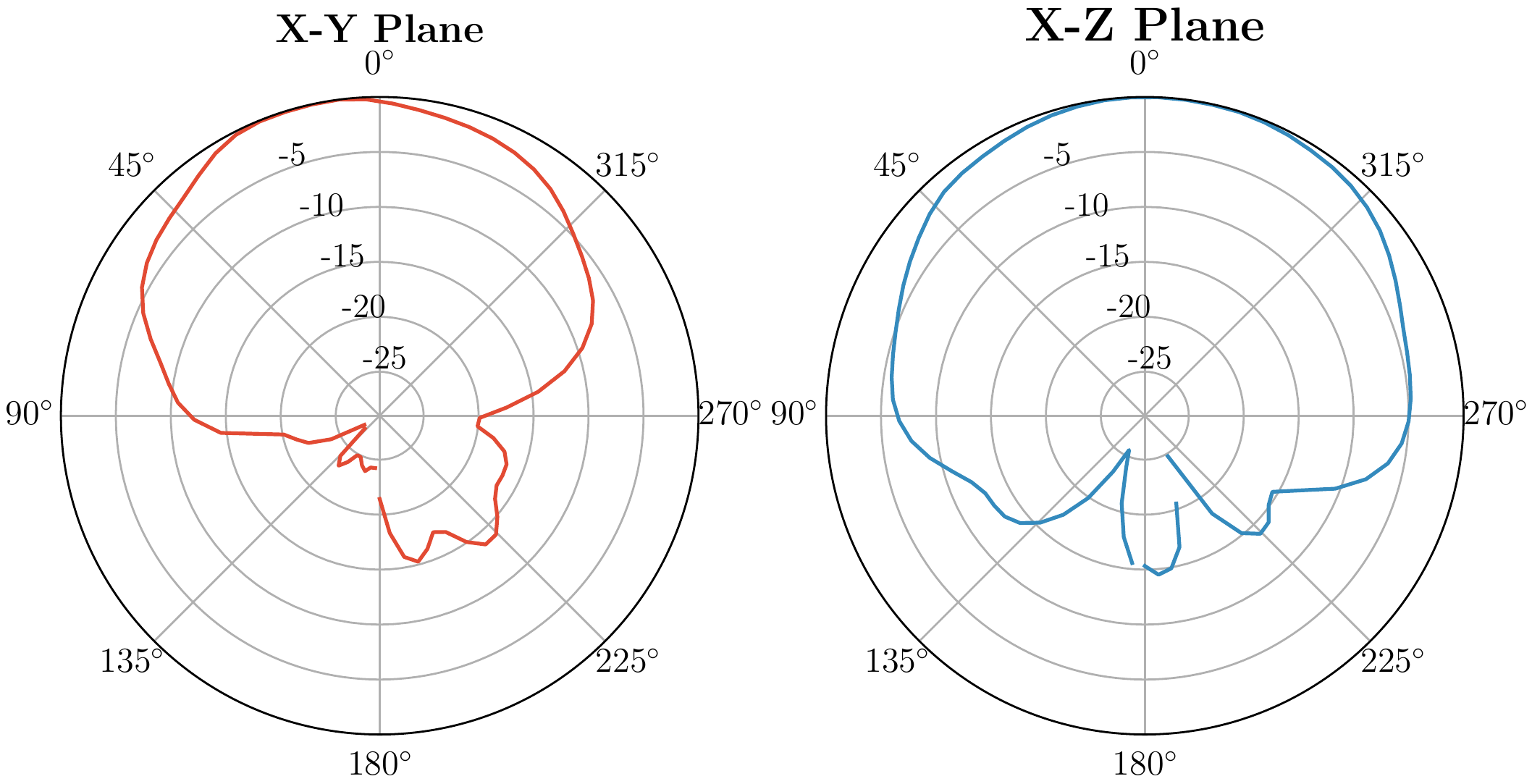}
\caption{Antenna radiation patterns in X-Y (horizontal) and X-Z (vertical) planes. In X-Y plane, half-power-beamwidth (HPBW) of roughly \(72^o\) is observed. }
\end{figure}
Fig 4 shows the radiation patterns of the surveillance antenna in X-Y (horizontal) and X-Z (vertical) planes. Since a single antenna with a half-power-beam-width (HPBW) of \(72^o\) in the horizontal plane is proposed, the system is not able to capture breathing patterns if the observer is out of that range. The HPBW is comparatively higher in the vertical plane indicating that the system is able to capture breathing variations even when the target's chest is elevated above or below the plane of surveillance antenna.

\section{Implementation}
\subsection{Dataset}

In order to train a deep learning model, a significant evaluation dataset is required to both train the model and validate results. An important consideration is the sample window size which presents a compromise between model accuracy and model complexity. A typical human adult exhales and inhales 12-18 times a minute. However, this range extends in the elderley and is typically between 10-30 breaths per minute. In order to capture sufficient signal variations and separate the breathing movement from random motions, a large window size is needed. This presents two challenges of increased model complexity and  training inconvenience. A compromise can be reached by selecting a window size of 10 seconds, which is large enough to capture 2-5 breaths per minute but small enough to conveniently train the model while keeping the complexity at a manageable level. 

The dataset\footnote{The complete dataset is available at http://ipt.seecs.nust.edu.pk/, along with relevant documentation and helper functions (in Matlab and Python).} consists of three activities corrsponding to breathing motion, static environment and random motion (such as human limb movement). For each activity, 480 samples of ten seconds each are recorded which equates to 240 minutes of training. 35\% of breathing data contain labels of the precise breathing rate during the activity, and is used for training the Random Forest model. Fig 5 shows the time and frequency domain (Fast Fourier Transorm) visualizations of the dataset. All three activities have distinct patterns in the time domain but the frequency domain patterns are not clearly apparent. The goal of the deep learning model is to learn the time domain representations and accurately classify the breathing activity.

\begin{figure}[!t]
\centering
\includegraphics[scale=0.28]{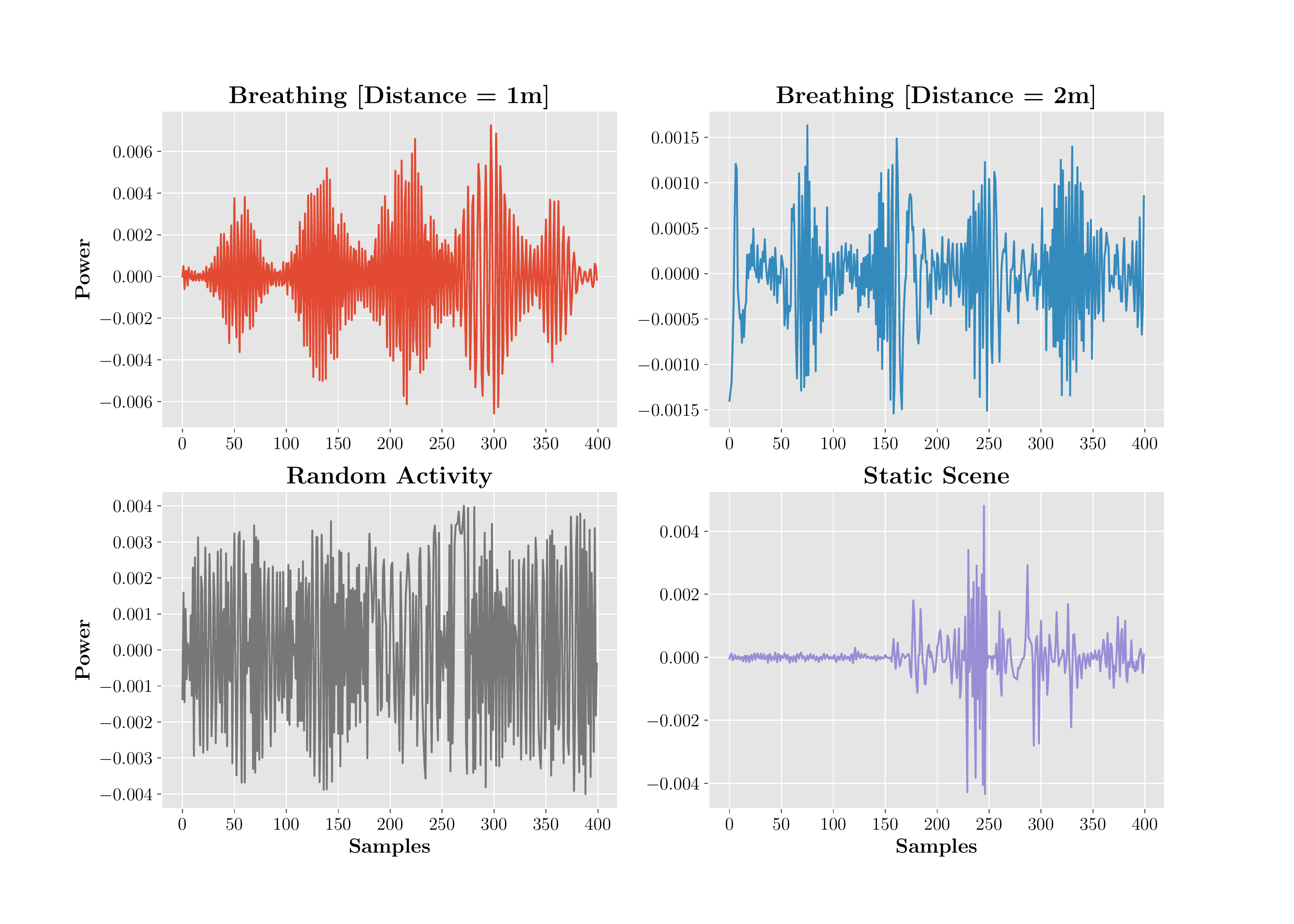}
\caption{Dataset time domain visualisations of (top) breathing activity at distances of 1 and 2 meters from the surveillance antenna, (bottom-left) random motion, and (bottom-right) static scene. }
\end{figure}

\subsection{Network Architecture}

The CNN is implemented on Keras which runs on top of TensorFlow and provides a modular approach to create deep networks. The CNN consists of a four layer model with two convolutional layers and two fully connected layers. For the first three layers, a rectified linear activation function (ReLU) is used which is defined as 

\begin{equation}
{f{(x)} = max(0, x)},
\end{equation}

\noindent where \(x\) is the input to the ReLU and \(max\) operators picks the argument with the maximum value. The final layer uses a softmax activation function to predict the output activity. A \(||W||_2\) norm penalty is added on the convolution layer weights as a regularization procedure to prevent data over-fitting [12]. To create sparse solutions, an \(||h||_1\) norm penalty is added on the activation layer [12]. After each layer, dropout is used to prevent the network from specializing on a single set of conditions. In addition, a pooling layer with filter size \(2 \times 2\) is used to progressively reduce the amount of parameters in the network to lessen computation resources and avoid overfitting.

The experimental setup is shown in fig 6. The passive sensing system used in the experiments utilizes USRP B200 software defined radio with an omni-directional antenna as an access point. The access point transmits orthogonal frequency division multiplexed symbols at a data rate of 3 Mb/s with a code rate of \(\frac{1}{2}\) and quadrature phase shift keying modulation. With this configuration, the transmit power of the WiFi source is estimated to be around -60 dBm. At the receiver end, we have log-periodic directional antennas with 6dBi gain and 60 degree beam-width as shown in Fig. 4 on surveillance antenna 1 and 2. The signals received at these antennas are digitized through a Spartan 6 XC6SLX75 FPGA and 61.44 MS/s, 12 bit ADC. 

\subsection{Hardware}
\begin{figure}[!t]
\centering
\includegraphics[scale=0.16]{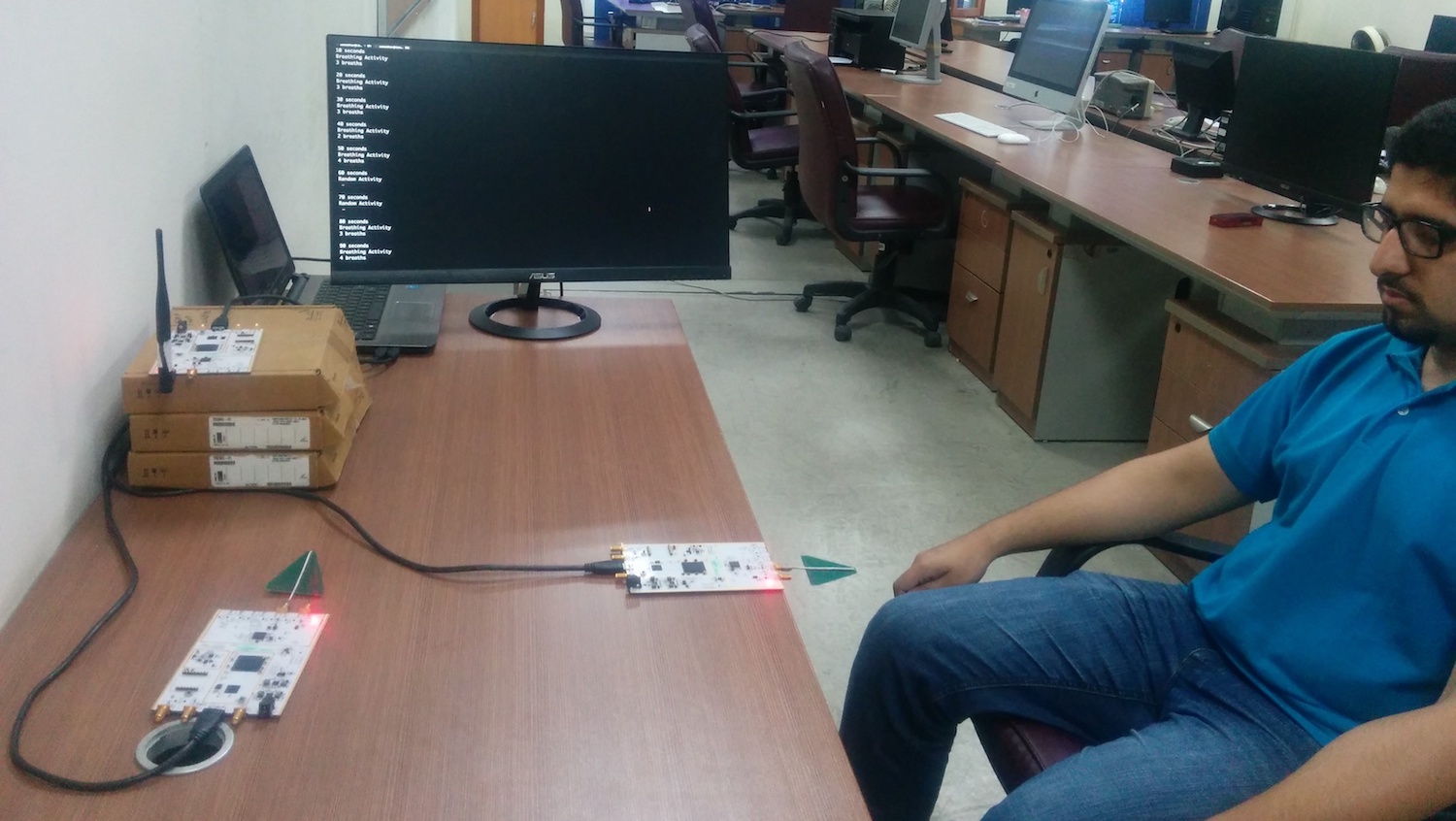}
\caption{Experimental setup with a target, surveillance and reference antennas, an access point, and computer display.}
\end{figure}

The training dataset is generated using GNU Radio and software defined radios at a sampling rate of 64 KS/s which is later resampled to 400 S/s. Afterwards, the time series signals is sliced up into a test and training dataset using a sliding window of ten seconds.

\subsection{Software}
We implement Python blocks in GNU Radio for deployment of machine learning modules. The CNN training is done on Keras which runs on top of Tensorflow and provides a modular framework to implement deep networks. The Random Forest classifier is trained using Python \textit{scikit-learn} library. The code runs in real-time and the display is updated on Ubuntu console every 2 seconds.

\subsection{Participants and Ground Truth}
In our experiments, we used a pulse oximeter to establish ground truth for breathing rate measurements. In order to increase data diversity, we employed 3 people with different fitness conditions and they all wore different outfits (sweaters, t-shirts et cetera).

\begin{figure}[!t]
\centering
\includegraphics[scale=0.31]{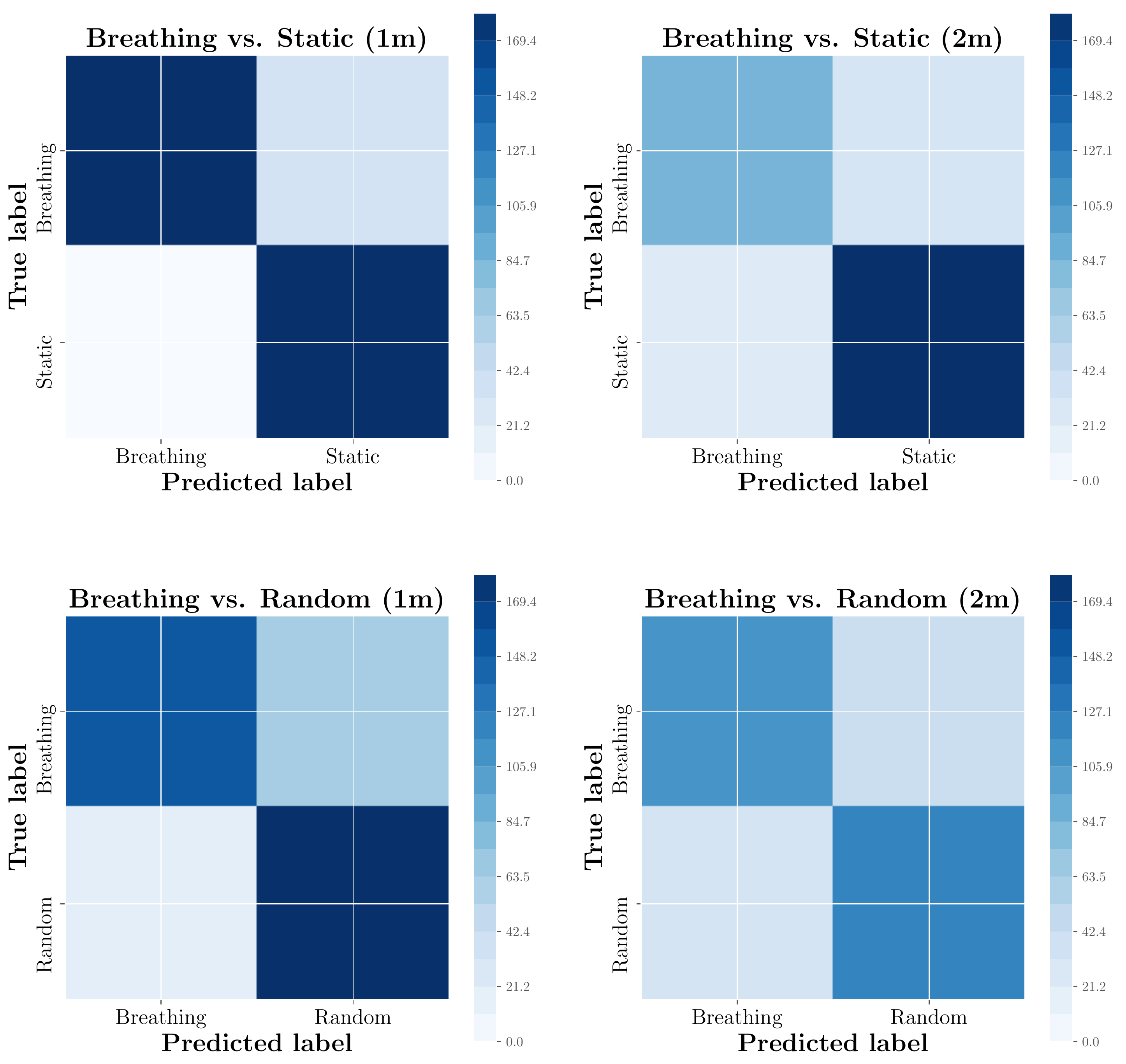}
\caption{Confusion matrices for deep CNN network trained to separate breathing activity from static and random scenes independently at distances of 1 and 2 meters from the surveillance antenna.}
\end{figure}

\begin{figure}[!t]
\centering
\includegraphics[scale=0.31]{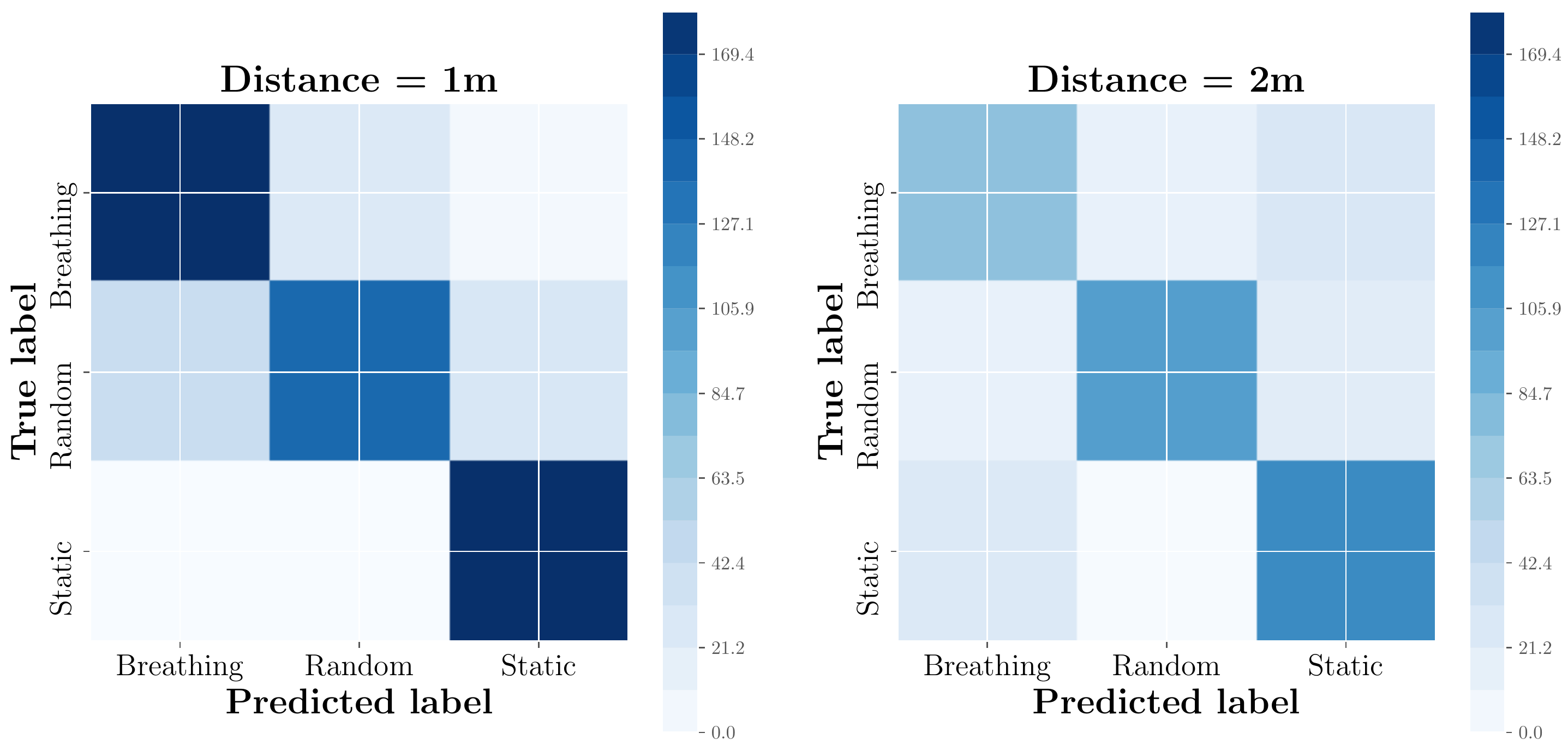}
\caption{Confusion matrices for deep CNN network trained to classify breathing, static and random activities simultaneously.}
\end{figure}

\section{Results}

This section provides the performance of the proposed system with respect to various parameters. Fig 7 shows the results of two CNN classifiers trained independently to classify breathing from static and random activities, respectively. At a distance of 1 meter from the surveillance antenna, the CNN classifier performs reasonably well giving an accuracy of 94.85\% (with a sensitivity of 100\% and specificity of 90.6\%) when classifiying breathing from a static scene and 81.03\% (with a sensitivity of 89.4\% and specificity of 75.49\%) when breathing is compared with a random scene. At a distance of 2 meters, the performance of these models drops to 74.26\% (69.29\% sensitivity and 78.62\% specificity) and 76.16\% (77.3\% sensitivity and 75.15\% specificity), respectively. 

These trends can be explained by the observation that as the diversity of data increases, more training samples are required to develop patterns and subsequently perform accurate classifications. At a lower distance, the SNR is high allowing the CNN model to accurately capture the variations in breathing activity. However, the SNR drops as the distance increases, requiring higher number of samples to develop clear patterns. Similarly, a random scene captures a wide variety of motions compared to a static scene in a controlled environment. These results are validated from the studies in computer vision domain, where the CNN model becomes increasingly robust with an increase in data samples and is able to detect diverse variations of the same image [20][21]. Fig 8 shows performance results when all activities are trained together. The trained CNN model achieves an accuracy of 85.17\% at distance of 1 meter and 74.55\% when the target is 2 meters away from the surveillance antenna.

While designing CNN models, the two most important parameters that need to be tuned are the number of convolution layers and the kernel depth in each convolution layer. Intutively, two convolution layers are sufficient for analyzing radio signals with the first layer capturing low level details such as edges in the signal, while the second layer classifying the activity. In order to determine optimized kernel depth, various iterations of CNN training are carried out. The results of this experiment are shown in fig 9. For two activities, the performance of the trained models does not significantly change with varying kernel depths. This is in contrast to the three activity model whose accuracy peaks at a kernel depth of 11 and drops before and afterwards. At a lesser depth, the model is not complex enough to capture signal variations and at a higher depth, the model overfits on the training data leading to a loss in validation set accuracy. This shows that as the complexity of the system increases, a more comprehensive model is required to capture a wide variety of variations.

\begin{figure}[!t]
\centering
\includegraphics[scale=0.5]{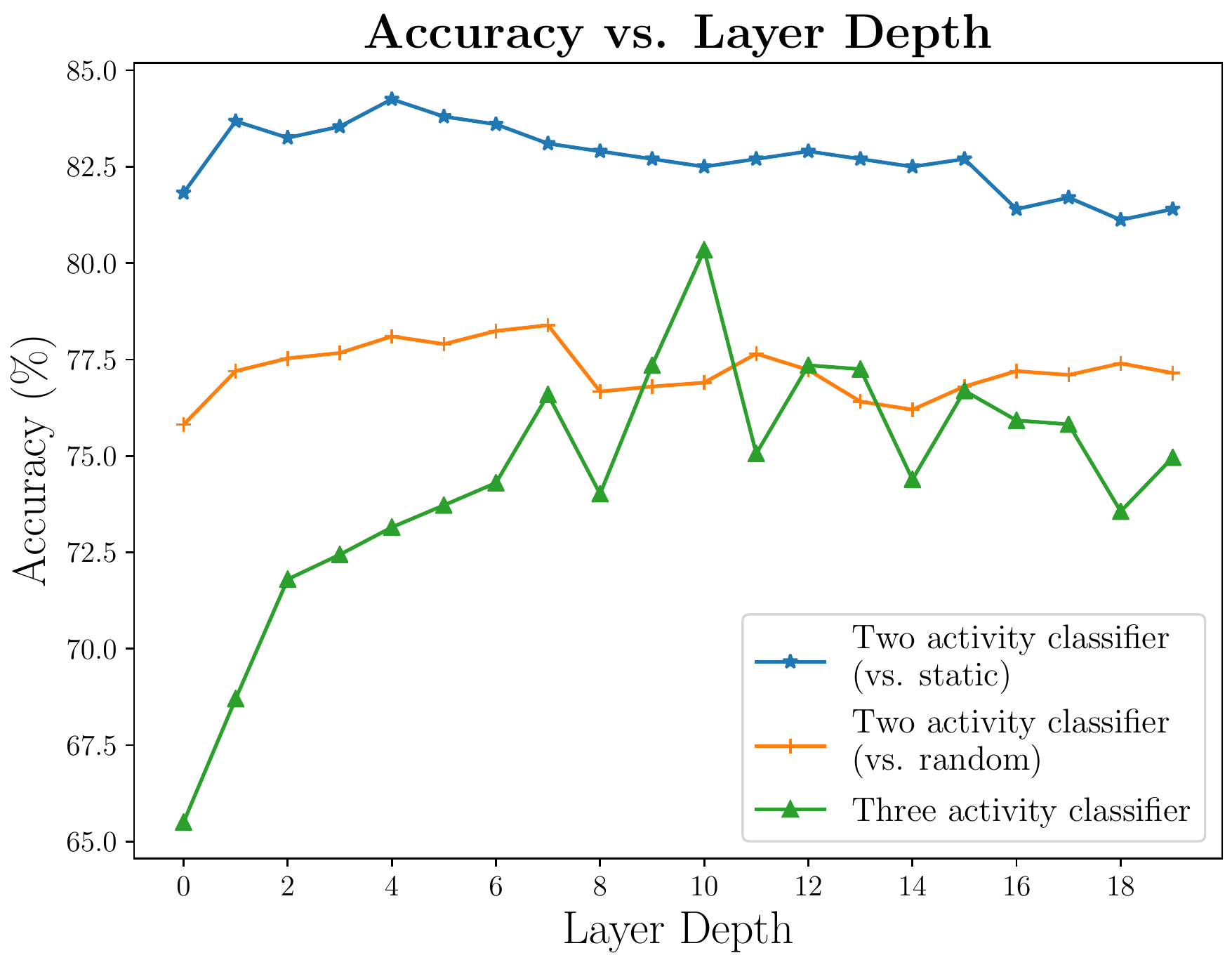}
\caption{Accuracy of CNN models with variations in kernel depth.}
\end{figure}

\begin{table}[ht]
\caption{Model Accuracy Vs. Samples} % title of Table
\centering % used for centering table
\begin{tabular}{c c c c c} % centered columns (4 columns)
\hline\hline %inserts double horizontal lines
Metric & Samples = 120 & Samples = 240 & Samples = 360 & Samples = 480 \\ [0.5ex] % inserts 
%heading
\hline % inserts single horizontal line
Accuracy & 68.15\% & 76.12\% & 78.28\% & 79.67\%  \\ % [1ex] adds vertical space
Increment & - & 7.97\% & 2.16 \% & 1.39\%  \\  
relative to\\
previous\\
reading\\[1ex] % [1ex] adds vertical space
\hline %inserts single line
\end{tabular}
\label{table:nonlin} % is used to refer this table in the text
\end{table}

Table I shows the accuracy of the deep activity classifier as the number of training data samples increases. When the training dataset consists of 40 samples, the classifier achieves an accuracy of 68.15\%. As the dataset size grows larger, the accuracy of the model increases while the accuracy differential decreases. For example, increasing data samples from 120 to 240 gives a performance rise of 7.97\% while an equivalent jump from 360 samples to 480 samples gives`' a performance imporovement of only 1.39\%. Accordingly, one may define a grade of service to determine an appropriate size of training dataset and thus avoid spending a big chunk of resources for minute performance improvements. For example, if the grade of service is set to 75\%, 250 samples may be sufficient in the above mentioned case. However, we note that the above analysis holds true only when the model is trained in a similar set of conditions. If the training environment is diverse, the CNN model will be more robust and performs well in new testing environments.

\begin{figure}[!t]
\centering
\includegraphics[scale=0.44]{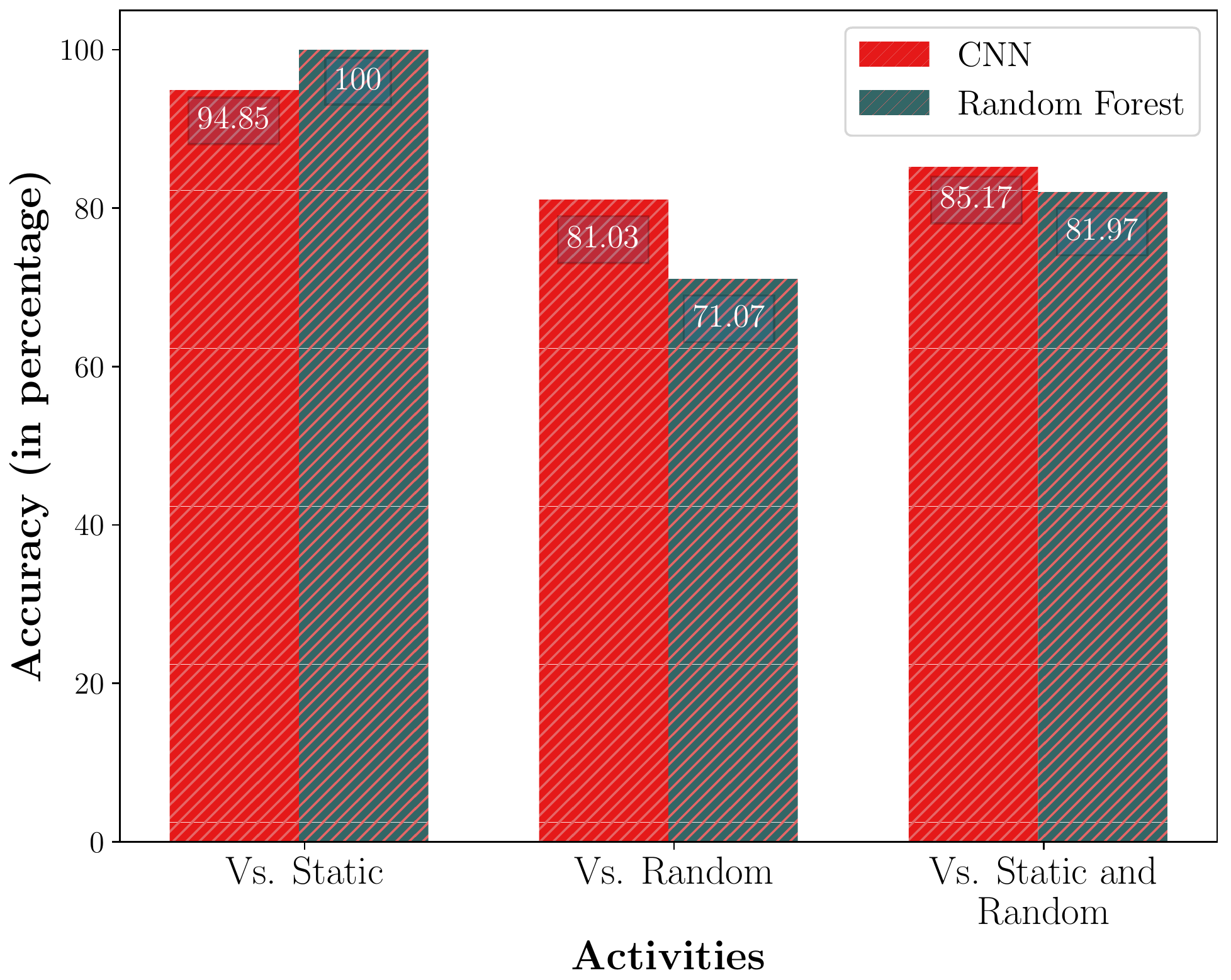}
\caption{Comparison of CNN and Random Forest models on various activities.}
\end{figure}

Next, we compare the performance of CNN model with Random Forest algorithm. In a simple two activity model where static and breathing activities are classified, Random Forest algorithm is highly robust and gives an accuracy of 100\% whereas the CNN model manages 94.85\%. However, as the classification problem gets complex, the CNN overtakes Random Forest algorithm in terms of validation set accuracy. When classifying breathing from random motion, CNN provides roughly 10\% improvement (81.03\% compared to 71.07\%). For the three activity model, CNN has an accuracy of 85.17\% while Random Forest is accurate for 81.97\% of the validation test cases.

\begin{figure}[!t]
\centering
\includegraphics[scale=0.52]{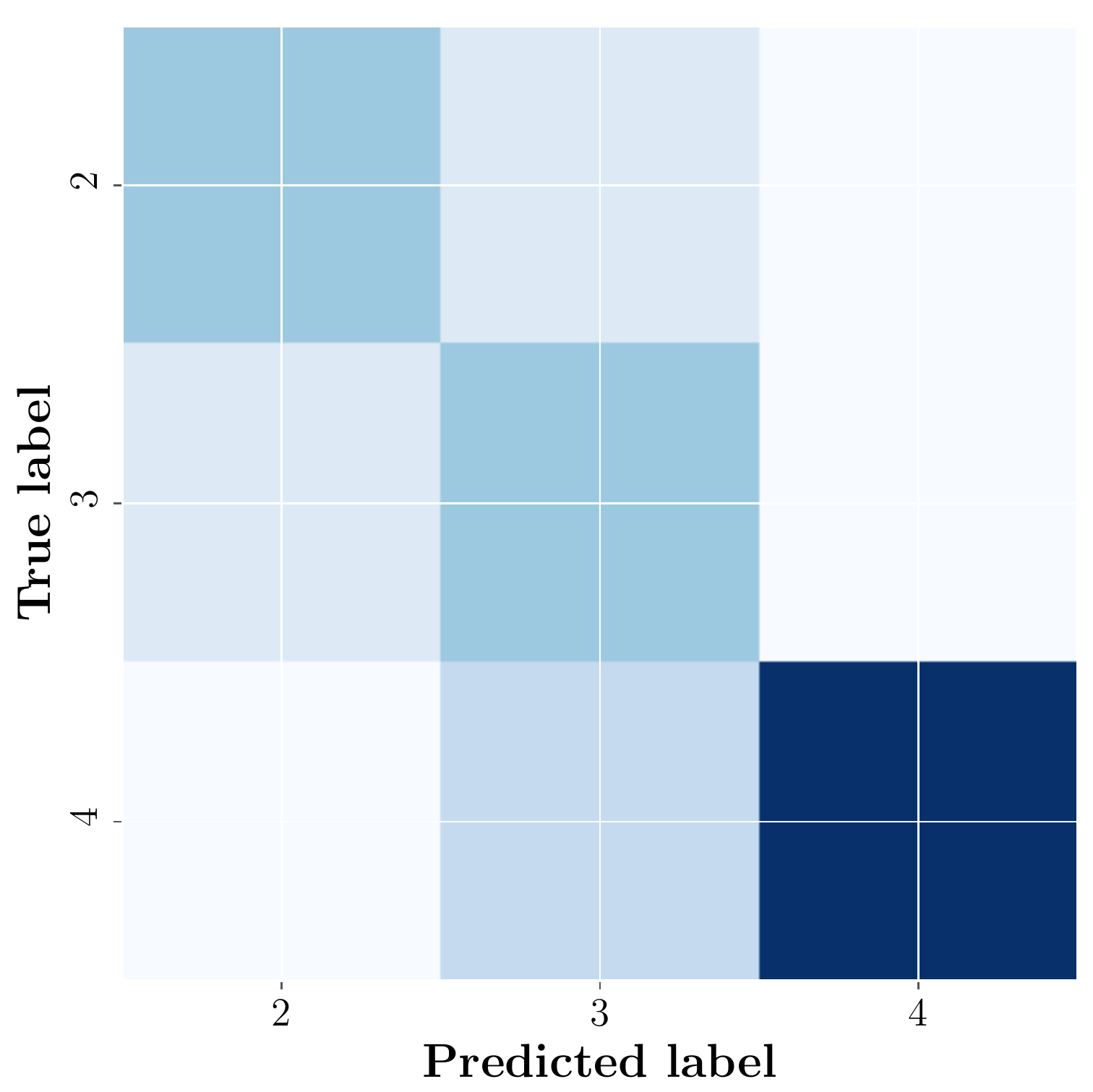}
\caption{Breathing Rate Estimation through Random Forest Classifier.}
\end{figure}

Finally, the performance of random forest classifier for the breathing estimation task is analyzed. The model gives an accuracy of 77.2\% in estimating human breaths. Due to the limited availability of labelled data, deep learning approach was not tested. In future, more training data can be added to gain better resolution and accuracy.

\subsection{Limitations and Future Work}
Notwithstanding the advantages of deep learning techniques, they require extensive data collection which may prove challenging for some applications. Realizing this challenge, the computer vision community has amassed a number of standard datasets in applications like autonomous driving, remote sensing, facial recognition, and surveillance. The radio community could benefit from a similar effort and this work is a step in that direction. Future work may focus on collecting respiration data in different environments and different body postures. A more ambitious attempt would be to build a deep CNN model to monitor human respiration during walking or running. Through extensive data collection and a comprehensive CNN model, this problem remains within reach.

Although the current research targeted only three activities, more can be added in the future with little modifications in the current model. For example, conditions like tremor (a condition in which human hand shakes) and human fall may be investigated. Beyond the health domain, the deep learning model proposed in this study can be extended to monitor a broad array of general human activities and gestures. The future is ripe with such possibilities.

\section{Conclusion}
In future, we envision smart homes capable of monitoring human activities and taking actions when required. Respiration is one of the key human activities, and may be used to monitor the stress levels and psychological states of a person. The past approaches that address this domain work well in a controlled environment but fail to detect when there are other motions involved. In addition, these techniques work poorly when the environments are changed or when new activities are added. This paper explores deep learning as an alternative approach to develop an end-to-end solution to monitor human respiration activity. We believe that deep learning marks a transition in the passive wireless sensing domain and may further be explored in the future research for a wide range of activities.

\end{document}